\pdfoutput=1

\documentclass[11pt]{article}

\usepackage{ACL}

\usepackage{times}
\usepackage{latexsym}

\usepackage[T1]{fontenc}

\usepackage[utf8]{inputenc}

\usepackage{microtype}

\usepackage{inconsolata}

\usepackage[utf8]{inputenc} %
\usepackage[T1]{fontenc}    %
\usepackage{hyperref}       %
\usepackage{url}            %
\usepackage{booktabs}       %
\usepackage{amsfonts}       %
\usepackage{nicefrac}       %
\usepackage{microtype}      %
\usepackage{xcolor}         %

\usepackage{colortbl}
\usepackage{tablefootnote}
\usepackage{graphicx}
\usepackage{multirow}
\usepackage{enumitem}
\usepackage{wrapfig}
\usepackage{tipa}
\usepackage[normalem]{ulem}

\newcommand{\p}[1]{{\flushleft \textbf{#1.}}}

\title{Citation: A Key to Building Responsible and Accountable \\ Large Language Models}

\author{Jie Huang $\quad$ Kevin Chen-Chuan Chang \\
Department of Computer Science, University of Illinois at Urbana-Champaign \\
 \texttt{\{jeffhj, kcchang\}@illinois.edu}
}

\begin{document}
\maketitle
\begin{abstract}

Large Language Models (LLMs) bring transformative benefits alongside unique challenges, including intellectual property (IP) and ethical concerns. This position paper explores a novel angle to mitigate these risks, drawing parallels between LLMs and established web systems. We identify ``citation''—the acknowledgement or reference to a source or evidence—as a crucial yet missing component in LLMs. Incorporating citation could enhance content transparency and verifiability, thereby confronting the IP and ethical issues in the deployment of LLMs. We further propose that a comprehensive citation mechanism for LLMs should account for both non-parametric and parametric content. Despite the complexity of implementing such a citation mechanism, along with the potential pitfalls, we advocate for its development. Building on this foundation, we outline several research problems in this area, aiming to guide future explorations towards building more responsible and accountable LLMs.
\end{abstract}

\section{Introduction}

The landscape of artificial intelligence is undergoing rapid transformation, spurred by the emergence of large language models (LLMs) such as ChatGPT/GPT-4~\citep{openai2022chatgpt,openai2023gpt4}. These models, recognized for their striking ability to generate human-like text, have shown enormous potential in various applications, from information provision to personalized assistance. Nonetheless, their capabilities bring along substantial risks, including intellectual property (IP) and ethical concerns~\citep{carlini2021extracting,carlini2023quantifying,huang2022large,shao2023quantifying,li2023multi,lee2023language,frye2022should,chesterman2023ai,10.1145/3442188.3445922,brown2022does,el2022sok}.

Research by \citet{carlini2021extracting,huang2022large}, for instance, reveals that LLMs are prone to memorizing extensive segments of their training data, including sensitive information. This can result in violations of IP and ethical guidelines. Furthermore, studies by \citet{el2022sok,brown2022does} suggest that current protective measures fail to provide a comprehensive and meaningful notion of safety for LLMs, making it seemingly impossible to develop safety-preserving, high-accuracy large language models even when trained on public corpora.

While the notion of building an entirely safe LLM might appear daunting, it is crucial to acknowledge that many well-established systems, such as the Web, grapple with similar challenges and have not yet reached absolute safety. Recent legislation like the Online News Act\footnote{\url{https://www.canada.ca/en/canadian-heritage/services/online-news.html}}, which requires online search engines to compensate Canadian online news outlets for their content, underscores the ongoing issues around content use and compensation on the Web. Furthermore, the Web continues to be a breeding ground for both sensitive information and misinformation. Hence, expecting a completely risk-free LLM may be an over-ask. Instead, our focus should be on accurately quantifying these risks and developing effective mitigation strategies. It is not about achieving absolute security, but about responsibly managing and minimizing risks in an ethically sound manner.

Guided by these insights, we propose to examine the problem through a different lens: \textit{Can we draw parallels between the risks inherent to LLMs and those experienced by established systems such as search engines and the Web?} \textit{Can we devise strategies to decrease these risks by aligning with the practices of these mature systems?}

In examining systems like the Web and search engines, we observe a common and robust practice employed to manage IP and ethical concerns: the use of ``citations''. Broadly defined, a ``citation'' refers to the act of mentioning or referencing a source or piece of evidence. For example, search engine results also serve as a form of citation, with each entry typically consisting of a title, URL, and brief description. These components collectively cite the webpage's content, offering the user an overview and inviting them to explore the source in greater depth. Citations thus act as anchors for accountability and credit in these systems, providing traceability, preventing plagiarism, and ensuring credit is correctly attributed. They also contribute to transparency, allowing users to verify the information's source.

Upon reflection, it becomes clear that LLMs lack this critical functionality. When LLMs generate content without citations, their output is perceived as independent and self-derived. This creates two significant issues: firstly, when the model produces valuable information, it fails to credit the source it relies on; secondly, when it generates harmful content, it becomes challenging to assign accountability. 
Incorporating the ability to cite could not only address these ethical and legal conundrums but also bolster the transparency, credibility, and overall integrity of the content generated by LLMs.

However, implementing a ``citation'' mechanism in LLMs is not as straightforward as it might seem. Unlike the Web, which explicitly links and references sources, LLMs internalize the information and transform it into hidden representations, making accurate citation a significant technical challenge.
Although some strides have been made in this direction, as seen in systems like New Bing\footnote{\url{https://www.bing.com/new}} and Perplexity AI\footnote{\url{https://www.perplexity.ai}}, they fall short on several fronts. First, the citations provided in the response of existing systems are often inaccurate \citep{liu2023evaluating,gao2023enabling}. Moreover, these systems typically only cite \textit{non-parametric} content, i.e., content directly retrieved from external sources such as the Web. However, they neglect \textit{parametric} content, the knowledge embedded in the model parameters, which also needs appropriate credit attribution and consideration for potential harm.

This position paper embarks on an exploratory journey into the potential of integrating a citation mechanism within large language models, examining its prospective benefits, the inherent technical obstacles, and foreseeable pitfalls. We delve into approaches to cite both non-parametric and parametric content, considering the unique characteristics of each type. We also identify and discuss potential setbacks, such as reduced creativity, dissemination of sensitive information, and citation bias. Building on this foundation, we lay bare the hurdles in our path, presenting them as enticing problems for future research. Through this endeavor, we aim to stimulate further discussion and research towards building responsible and accountable large language models.

\section{Overview of Large Language Models}

Large language models are typically built on the foundation of transformer architectures~\citep{vaswani2017attention}.
The training process of LLMs usually involves self-supervised learning on vast quantities of text data, including books, articles, and internet content, primarily sourced from the Web. During this stage, models are exposed to diverse textual data, allowing them to learn grammar, facts~\citep{petroni2019language}, and even reasoning abilities~\citep{wei2022chain,huang2022towards}.

Following the initial training, models may undergo further training on smaller, labeled datasets.
For instance, ChatGPT~\citep{openai2022chatgpt}, a prominent LLM, is fine-tuned on a carefully curated dataset consisting of demonstrations and comparisons, which help the model learn how to generate appropriate responses in conversational contexts. 

\p{Risks in LLMs} 
While LLMs offer numerous benefits, they also pose significant  risks~\citep{carlini2021extracting,huang2022large,li2023multi,el2022sok,guo2022threats}. \citet{el2022sok} highlight these risks, concluding that it is fundamentally impossible to develop safety-preserving, high-accuracy LLMs due to the fundamental intrinsic impossibility of the foundation model learning problem. As they summarized, LLMs achieve optimal performance by employing high-dimensional interpolation, necessitating vast quantities of user-generated data. However, language data from genuine users is intrinsically diverse, with significant variations in individual preferences and styles. 
This results in empirical heterogeneity, which in turn contributes to the vulnerability of LLMs, particularly when handling sensitive data or encountering fabricated information from fake accounts.

\section{``Citation'' in LLMs}

\begin{figure}[h]
\vspace{1mm}
\centerline{\includegraphics[width=\linewidth]{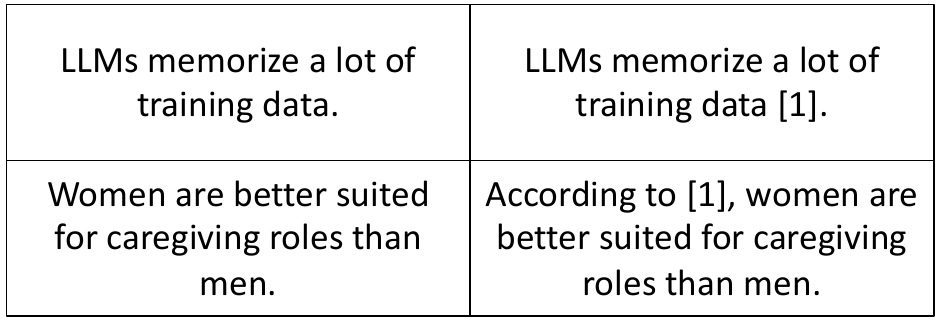}}
\caption{Examples without (left) and with (right) citations. In the first case, citations serve as a way to appropriately credit authors. In the second case, citing the original source of a biased statement ensures that the bias is not misconstrued as the model's opinion.}
\vspace{-2mm}
\label{fig:citation}
\end{figure}

As discussed in the introduction, expecting a risk-free LLM may be an over-ask. The key lies in responsibly managing and minimizing risks in an ethically sound manner. By drawing a comparison between LLMs and the Web, we find that ``citation'' is a key missing component in LLMs.

Figure~\ref{fig:citation} illustrates model generations with and without citations. In the absence of a citation, there is a potential risk of misunderstanding, leading one to believe that the claim is an opinion or statement formulated by the LLM itself. This not only fails to appropriately credit the original authors, but could also result in ethical dilemmas if the claim is inaccurate or misrepresented.

On the other hand, the inclusion of citations can act as a multifaceted solution to these concerns. Primarily, it helps to mitigate intellectual property and ethical disputes by signaling that the information is not a product of the LLM's ``opinion'', but a reflection of the cited source. Additionally, citations would enhance the transparency and verifiability of the LLM's output.
By indicating the source from which the information is derived, they provide a clear pathway for users to independently verify the validity and context of the information.\footnote{However, citation may also lead to certain potential pitfalls; please refer to Section~\ref{sec:pitfalls} and Section~\ref{sec:research_problems} for more details.}

\section{RoadMap}

In this section, we embark on exploring the potential of incorporating a ``citation'' mechanism within LLMs.
We start our exploration by defining when it would be ideal for an LLM to provide a citation, drawing insights from established practices in academia and existing systems like search engines and the Web. We then delve into discussing the possible strategies for effectively implementing citations in LLMs, confronting the methodological and technical intricacies this endeavor involves.

\begin{figure*}[tp]
\centerline{\includegraphics[width=\linewidth]{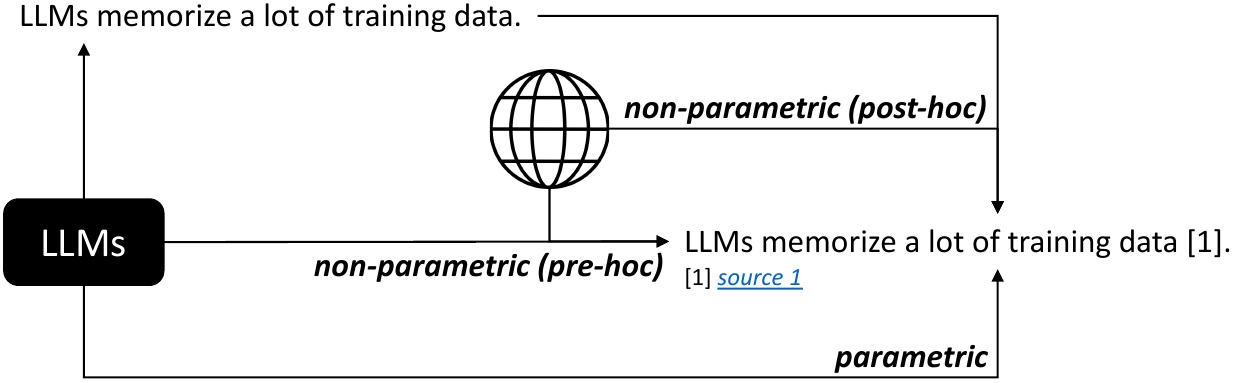}}
\caption{\textit{Non-parametric} and \textit{parametric} citations.}
\vspace{-1mm}
\label{fig:method}
\end{figure*}

\subsection{When to Cite?}

\label{sec:when_to_cite}

In academic or professional writing, a citation is typically required when using someone else's ideas, concepts, data, or specific language. For LLMs, determining when to provide a citation is a considerably more challenging task. Given the vast and varied range of queries posed to LLMs, it is crucial to establish when a citation would be appropriate or necessary.

A fundamental rule could be that any fact, idea, or concept that is not general knowledge should be cited. This mirrors the existing conventions on the Web, where sources for specific information are typically provided. For instance, widely known facts like ``The Earth revolves around the Sun'' would not necessitate a citation, while a less well-known fact like ``The fastest spinning stars can rotate more than 600 times per second'' would warrant one.

Moreover, the need for a citation could also depend on the nature of the task LLMs are performing. Certain tasks may not necessitate citations, particularly if the output is a reformulation or reinterpretation of the input. For example, in summarization tasks, LLMs condense the input data without introducing new information. The resultant summary is hence an interpretation of the input, and typically, a citation may not be needed for such tasks.
Similarly, translation tasks involve converting content from one language to another, without the introduction of novel information.

In essence, while the need for citations in LLMs is task-dependent and context-specific, the guiding principles should be the commitment to knowledge integrity, respect for intellectual property, and adherence to ethical norms.
These are similar principles that guide the management of intellectual property and ethical concerns on the Web and in search engines.

\subsection{How to Cite?}

\label{sec:how_to_cite}

Incorporating citations in LLMs ideally involves connecting outputs to the original source of information. However, this presents a notable technical challenge. During LLMs' training, information is transformed into hidden representations, unlike search engines which possess indices to track and retrieve information. In the case of LLMs, this index is absent, which makes referencing the original source a daunting task. In this section, we delve into the consideration of citations for both \textit{non-parametric} and \textit{parametric} content (Figure~\ref{fig:method}).

\subsubsection{Citation for \textit{non-parametric} content}
As a potential solution to prevailing challenges, one could design a hybrid system that merges large language models with information retrieval (IR) systems. In this approach, the model is trained to discern when a citation might be required. Subsequently, the IR system is utilized to retrieve relevant sources, namely, \textit{non-parametric} content. The LLM can then incorporate these sources into its responses as citations. We identify two strategies for citing non-parametric content:

\begin{itemize}[leftmargin=*, nolistsep, nolistsep, topsep=1mm] \setlength{\itemsep}{1mm}
\item \textbf{\textit{Pre-hoc citation}}: This approach involves first identifying the need for a citation in the upcoming dialogue or content generation. Once this requirement is recognized, the LLM triggers the IR system to retrieve the necessary information. The LLM then generates its response, seamlessly incorporating the retrieved non-parametric content as citations. This technique can be associated with the broader body of research that augments language models with retrieval \citep{pmlr-v119-guu20a,NEURIPS2020_6b493230,izacard-grave-2021-leveraging,pmlr-v162-borgeaud22a,izacard2022few,shi2023replug,wang2023shall,menick2022teaching,huang2023raven}.

\item \textbf{\textit{Post-hoc citation}}: Conversely, in this strategy, the LLM initially produces a response. An evaluation process then scrutinizes the generated content to ascertain whether a citation is necessary. If a citation is deemed necessary, the IR system is used to locate the appropriate non-parametric content, which is subsequently inserted into the existing text as a citation. Related research includes measuring or requiring attribution in LLMs~\citep{rashkin2023measuring,gao2022rarr,honovich2022true,yue2023automatic,liu2023evaluating,gao2023enabling}.
\end{itemize}

In practical applications, a combination of both \textit{pre-hoc} and \textit{post-hoc} citation methods could be adopted for an optimized method. This mixed approach would employ the initial identification and retrieval of potential citations in line with the \textit{pre-hoc} method, followed by a \textit{post-hoc} evaluation to refine the integration of citations based on the generated content. This blend of proactive retrieval and reactive refinement could facilitate the creation of robust, accurate, and well-supported content, while also mitigating intellectual property and ethical concerns surrounding LLMs.

\subsubsection{Citation for \textit{parametric} content}
\label{sec:parametric_citation}

In addition to the \textit{non-parametric} content, i.e., content directly retrieved from external sources such as the Web, \textit{parametric} content, which refers to information internalized from the training data, also needs appropriate credit attribution and consideration for potential harm. However, crafting a citation strategy for \textit{parametric} content presents its own set of unique challenges.

The fundamental challenge is the underlying nature of how LLMs process and internalize information. During training, LLMs assimilate vast amounts of data and transform them into an intricate, high-dimensional space that represents learned patterns and structures. The transformation process, rooted in complex mathematical operations, does not inherently retain any clear mapping back to individual data points in the training set. Consequently, generated content cannot easily be traced back to specific training data \citep{pmlr-v70-koh17a,NEURIPS2022_7234e0c3,park2023trak,grosse2023studying}.

This situation is further complicated by the fact that an output generated by LLMs is typically influenced by a multitude of training data points, rather than a single source. This is due to the multi-faceted and context-sensitive nature of language understanding and generation, where a single output can be influenced by a diverse range of linguistic patterns and structures. Thus, the task of accurately attributing a generated output to specific training data pieces is a complex and multifaceted problem that involves unpacking the high-dimensional representations in the model.

Despite these challenges, potential solutions exist. A conceivable approach involves \textit{\textbf{training the model with source identifiers}}, essentially tags that link specific pieces of information back to their original sources.  During training, the model could then be encouraged to retain these identifiers.
This would provide a more transparent lineage of information, thereby enhancing accountability. 
A relevant attempt in this direction was made by \citet{taylor2022galactica}, which used special reference tokens to wrap citations and trained models to predict these citations. However, it exhibited certain limitations, such as citation inaccuracy and confinement to academic citations. 
The successful execution of this method would likely call for advancements in model architecture and training techniques, thereby highlighting intriguing directions for future research.

\section{Pitfalls of Citation in LLMs}

\label{sec:pitfalls}

While citations in LLMs can potentially mitigate risks such as IP and ethical issues, as well as improve transparency and verifiability, it is crucial to consider potential pitfalls.

\p{Over-Citation and Sensitive Information Dissemination}
The implementation of a citation system in LLMs poses the risk of over-citation, where the excessive use of references might expose more information than necessary. This overexposure could lead to information overload, diluting the significance of critical citations. Moreover, over-citation might inadvertently elevate the risk of disseminating sensitive information \cite{huang2022large,shao2023quantifying,li2023multi}. An ill-intentioned user could exploit these extensive citations to gather additional sensitive information.

\p{Inaccurate Citations}
Another potential pitfall of implementing citations in LLMs is the risk of inaccurate citations \citep{liu2023evaluating,gao2023enabling}. Given that LLMs may not possess a deep understanding of the content they are trained on or the sources they are citing, there is a chance that they could incorrectly attribute information to a source that does not actually contain that information. Inaccurate citations could mislead users, causing them to believe that a piece of information is verified and supported by a credible source, when in fact, it is not. 

\p{Outdated Citations}
With the continuous expansion and evolution of knowledge, there is a risk that the sources an LLM cites may become outdated or irrelevant over time. This is particularly likely in fast-evolving fields where new discoveries or advancements quickly supersede existing knowledge. As LLMs are trained on a fixed dataset, their generated content and the sources they cite may not reflect the most current or accurate information. Therefore, there is a potential for LLMs to propagate outdated knowledge, misleading users who rely on the generated content and the cited sources for information.

\p{Propagation of Misinformation}
The risk of propagation of misinformation presents a significant concern in the application of LLMs \citep{pan2023risk}. As LLMs generate output based on the data they have been trained on, there is a chance they could inadvertently cite or echo unreliable or misleading sources, thereby spreading misinformation. This problem could potentially be amplified by the addition of a citation mechanism. A misinterpreted or incorrect citation could be perceived as an authoritative endorsement, inadvertently lending credibility to inaccurate or misleading content. 

\p{Citation Bias}
Implementing citations in LLMs can also lead to citation bias~\citep{jannot2013citation,greenberg2009citation,10.1145/3442188.3445922,metzler2021rethinking,10.1145/3498366.3505816}. Models may tend to cite certain types of sources over others, either due to the characteristics of the training data or inherent biases in the retrieval mechanism of the IR system. This could lead to an over-reliance on certain types of information and unintentional promotion of certain viewpoints.

\begin{figure*}[tp]
\centerline{\includegraphics[width=\linewidth]{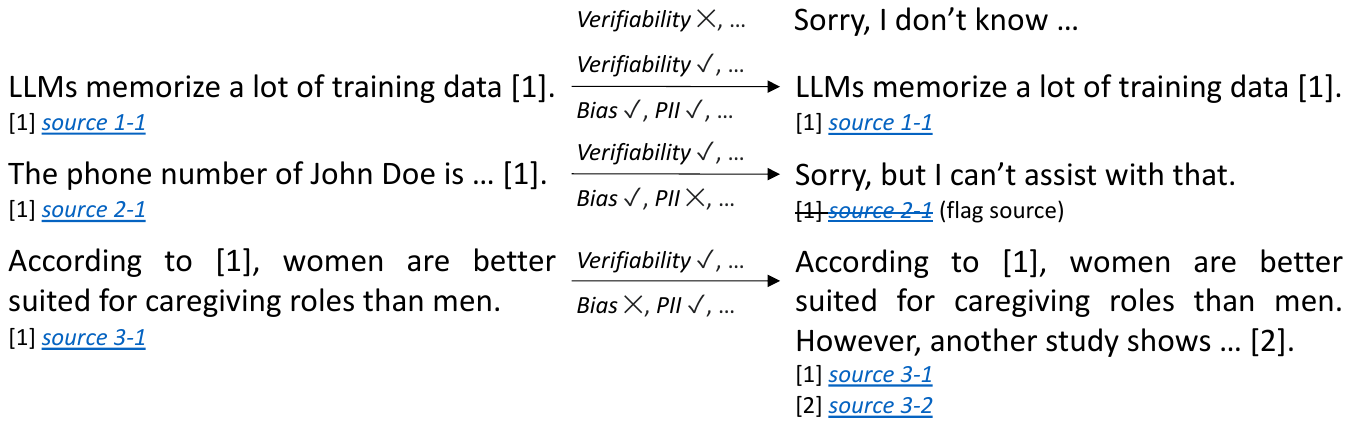}}
\caption{Citation with a multifaceted implementation. 1) If a statement cannot be verified by a reliable source, the model can learn to respond with ``I don't know''; 2) If the generated output contains sensitive information, such as Personally Identifiable Information (PII), the model should refuse to answer and flag the source to alert the maintainer; 3) If the generated output is detected to reflect a certain bias, the model should refine its response to be more comprehensive and unbiased.}
\vspace{-1mm}
\label{fig:approach}
\end{figure*}

\p{Potential for Diminished Creativity}
The integration of citations could inadvertently cause a decrease in the creative outputs of the model. When prompted to generate innovative text or propose creative solutions, LLMs might become over-reliant on existing, citable information, thus stifling their novel content generation.

\p{Legal Implications}
The utilization of citations could also bring forth legal implications. The introduction of citations could imply a level of responsibility and accountability that the LLM, as an artificial entity, is not equipped to handle. Legal systems around the globe have not yet achieved a consensus on addressing legal issues associated with artificial intelligence, its outputs, and the individuals or entities that create and operate these systems. The inclusion of citations could further complicate these discussions.

\section{Barriers and Research Problems}
\label{sec:research_problems}

Building on the potential solutions and pitfalls discussed above, we delve into the primary barriers and corresponding research problems that need to be addressed for successful citation implementation in large language models. Figure~\ref{fig:approach} illustrates examples showing that the inclusion of a citation should be combined with a multifaceted implementation by addressing these research problems.

\subsection{Determining When to Cite}

Deciding when an LLM should cite its sources is a complex issue. While it may be intuitive to suggest that LLMs should always cite sources when they generate information that is not common knowledge (\S\ref{sec:when_to_cite}), defining what constitutes ``common knowledge'' is itself a difficult task. Furthermore, as discussed in \S\ref{sec:pitfalls}, it is essential to consider the potential risks associated with over-citation, particularly the increased risk of sensitive information dissemination~\cite{huang2022large,shao2023quantifying,li2023multi}. LLMs may inadvertently expose sensitive information or contribute to information overload if they include unnecessary or excessive citations. Balancing the need for transparency and accountability with the need to protect privacy and prevent information overload is a critical challenge that needs to be addressed.

\subsection{Addressing Hallucination in Citation}

Hallucination in large language models refers to the phenomenon where the models generate information not grounded in their training data, and that cannot be verified or is simply incorrect~\citep{ji2023survey,zheng2023does}. The incorporation of a citation feature can both alleviate and exacerbate this issue. On the one hand, requiring LLMs to link generated information to a tangible source can serve as a form of external verification, potentially restraining the model from generating completely baseless or hallucinated content. The requirement for a source may encourage the model to better align its output with the available data, thereby reducing the likelihood of hallucination.

On the other hand, the citation mechanism itself can potentially hallucinate. If not meticulously designed and implemented, it may end up citing incorrect or non-existent sources \citep{liu2023evaluating,gao2023enabling}. This presents a twofold challenge: Not only is the generated content incorrect, but the citation misleads users into believing that the content is verified and substantiated by the cited source. This issue necessitates the development of techniques to enhance the model's ability to accurately represent the information present in the source, and equip the model to cross-check the consistency of the generated content with the content of the cited source.

\subsection{Maintaining Temporal Relevance of Citations}

In the pursuit of an effective citation mechanism within LLMs, it is essential to address the need for the model's ability to stay updated with the most recent and relevant knowledge. 

One potential approach towards this challenge is inspired by the operational principles of search engines. In their bid to stay relevant, search engines continuously update their indexes and ranking algorithms to reflect the latest web. A similar approach could be adopted for LLMs, where they could be designed for ongoing training on updated datasets.

However, executing this in practice presents a significant research problem, considering the scale and complexity of continuously training LLMs and updating their citation mechanisms. Exploring efficient techniques for model training and designing citation mechanisms capable of consistently prioritizing the most recent and relevant sources will require substantial research and development. 

\subsection{Evaluating Source Reliability}

Another important challenge is evaluating the reliability of sources used for training data and citations. As mentioned in \S\ref{sec:pitfalls}, LLMs could potentially propagate misinformation if they cite unreliable or misleading sources. While search engines face similar challenges, they are equipped with advanced algorithms to evaluate the reliability and relevance of web pages~\citep{page1999pagerank}. Implementing analogous systems within the framework of LLMs presents an interesting and crucial direction for further exploration.

\subsection{Mitigating Citation Bias}

Citation bias in LLMs, as discussed in \S\ref{sec:pitfalls}, can result in the uneven representation of information, leading to the propagation of certain viewpoints while others are neglected. Formulating strategies to curtail such tendencies is paramount.

To begin with, sourcing a more balanced selection of training data can mitigate bias at the inception stage. Ensuring diversity in terms of viewpoints and topics in the training data can reduce bias to some extent.

During citation retrieval, LLMs should utilize an impartial mechanism that does not favor specific types of sources. The underlying algorithms should be optimized to retrieve citations based on their relevance and credibility rather than the prominence of the source or its frequency in the training data.

Finally, the development and application of effective evaluation techniques can help identify and measure any residual bias in LLM outputs. Quantifying the extent of bias enables more targeted corrective measures and provides an objective measure of their efficacy.

\subsection{Balancing Existing Content with Novel Content Generation}

Another intriguing area of research centers on striking a balance between the frequency of citing existing content and generating novel content. LLMs are admired for their capacity to generate creative and unique content \citep{franceschelli2023creativity}, as well as their reasoning ability~\citep{wei2022chain,huang2022towards}. An over-reliance on citations could potentially inhibit these attributes, reducing the model to a mere aggregator of existing knowledge rather than a generator of new ideas.

Research into this would involve the development of techniques that allow for appropriate citation without hampering the model's creativity.
One potential approach could be to create models that are capable of determining the novelty of their generated content and adjusting their citation behavior accordingly. For instance, if a model is generating content based heavily on its training data or the retrieved content, it should provide appropriate citations. Conversely, if the model is generating content that is significantly different from its training data and the retrieved content, it might deem citation unnecessary. Developing such capabilities would require significant advancements in understanding how LLMs generate novel content and how to quantify the `novelty' of such content.

\subsection{Navigating Copyright and Fair Use Laws}

The application of citation mechanisms in LLMs opens up a new array of legal challenges. Understanding and complying with copyright and fair use laws when citing sources is a complex issue. For instance, how much quoted material from a source would be considered fair use and under what conditions can it be used? In many jurisdictions, the law is not completely clear, especially as it applies to the use of AI technology. Thus, research in the legal aspects of using LLMs for generating text with citations is crucial to ensure legal compliance.

\section{Conclusion}

In conclusion, the incorporation of a citation mechanism within LLMs presents a promising approach to numerous challenges, including but not limited to intellectual property rights, ethical concerns, and the need for transparency and verifiability in AI outputs. By equipping LLMs with the ability to accurately attribute the origins of information, we can cultivate a climate of enhanced accountability for the content these models generate. This signifies a progressive step towards constructing a framework of ethical responsibility in AI that respects intellectual property rights and upholds information integrity.
Through these efforts, we aim to foster more responsible, accountable, and reliable AI systems, ultimately contributing to a better, more trustworthy technological future.

\section*{Limitations}

While introducing a citation mechanism in LLMs presents an exciting opportunity for enhancing responsibility and accountability, implementing such a system is not without its technical challenges. Our paper introduces this concept with a hopeful perspective, but readers should be cognizant of the numerous technical hurdles that must be overcome, as highlighted in Section~\ref{sec:pitfalls} and Section~\ref{sec:research_problems}. Nevertheless, these challenges also represent valuable areas for future research and innovation. By addressing these issues head-on, we believe there is potential to unlock the true benefits of such a mechanism, leading to more responsible and accountable large language models.

\section*{Acknowledgements}

We thank the reviewers for their constructive feedback. This material is based upon work supported by the National Science Foundation IIS 16-19302 and IIS 16-33755, Zhejiang University ZJU Research 083650, IBM-Illinois Center for Cognitive Computing Systems Research (C3SR) and IBM-Illinois Discovery Accelerator Institute (IIDAI), grants from eBay and Microsoft Azure, UIUC OVCR CCIL Planning Grant 434S34, UIUC CSBS Small Grant 434C8U, and UIUC New Frontiers Initiative. Any opinions, findings, conclusions, or recommendations expressed in this publication are those of the author(s) and do not necessarily reflect the views of the funding agencies.

\bibliography{anthology,custom}

\begin{thebibliography}{47}
\expandafter\ifx\csname natexlab\endcsname\relax\def\natexlab#1{#1}\fi

\bibitem[{Bae et~al.(2022)Bae, Ng, Lo, Ghassemi, and Grosse}]{NEURIPS2022_7234e0c3}
Juhan Bae, Nathan Ng, Alston Lo, Marzyeh Ghassemi, and Roger~B Grosse. 2022.
\newblock \href {https://proceedings.neurips.cc/paper_files/paper/2022/file/7234e0c36fdbcb23e7bd56b68838999b-Paper-Conference.pdf} {If influence functions are the answer, then what is the question?}
\newblock In \emph{Advances in Neural Information Processing Systems}, volume~35, pages 17953--17967. Curran Associates, Inc.

\bibitem[{Bender et~al.(2021)Bender, Gebru, McMillan-Major, and Shmitchell}]{10.1145/3442188.3445922}
Emily~M. Bender, Timnit Gebru, Angelina McMillan-Major, and Shmargaret Shmitchell. 2021.
\newblock \href {https://doi.org/10.1145/3442188.3445922} {On the dangers of stochastic parrots: Can language models be too big?}
\newblock In \emph{Proceedings of the 2021 ACM Conference on Fairness, Accountability, and Transparency}, FAccT '21, page 610–623, New York, NY, USA. Association for Computing Machinery.

\bibitem[{Borgeaud et~al.(2022)Borgeaud, Mensch, Hoffmann, Cai, Rutherford, Millican, van~den Driessche, Lespiau, Damoc, Clark, de~Las~Casas, Guy, Menick, Ring, Hennigan, Huang, Maggiore, Jones, Cassirer, Brock, Paganini, Irving, Vinyals, Osindero, Simonyan, Rae, Elsen, and Sifre}]{pmlr-v162-borgeaud22a}
Sebastian Borgeaud, Arthur Mensch, Jordan Hoffmann, Trevor Cai, Eliza Rutherford, Katie Millican, George van~den Driessche, Jean{-}Baptiste Lespiau, Bogdan Damoc, Aidan Clark, Diego de~Las~Casas, Aurelia Guy, Jacob Menick, Roman Ring, Tom Hennigan, Saffron Huang, Loren Maggiore, Chris Jones, Albin Cassirer, Andy Brock, Michela Paganini, Geoffrey Irving, Oriol Vinyals, Simon Osindero, Karen Simonyan, Jack~W. Rae, Erich Elsen, and Laurent Sifre. 2022.
\newblock \href {https://proceedings.mlr.press/v162/borgeaud22a.html} {Improving language models by retrieving from trillions of tokens}.
\newblock In \emph{International Conference on Machine Learning, {ICML} 2022, 17-23 July 2022, Baltimore, Maryland, {USA}}, volume 162 of \emph{Proceedings of Machine Learning Research}, pages 2206--2240. {PMLR}.

\bibitem[{Brown et~al.(2022)Brown, Lee, Mireshghallah, Shokri, and Tram{\`e}r}]{brown2022does}
Hannah Brown, Katherine Lee, Fatemehsadat Mireshghallah, Reza Shokri, and Florian Tram{\`e}r. 2022.
\newblock \href {https://dl.acm.org/doi/fullHtml/10.1145/3531146.3534642} {What does it mean for a language model to preserve privacy?}
\newblock In \emph{2022 ACM Conference on Fairness, Accountability, and Transparency}, pages 2280--2292.

\bibitem[{Carlini et~al.(2023)Carlini, Ippolito, Jagielski, Lee, Tramer, and Zhang}]{carlini2023quantifying}
Nicholas Carlini, Daphne Ippolito, Matthew Jagielski, Katherine Lee, Florian Tramer, and Chiyuan Zhang. 2023.
\newblock \href {https://openreview.net/forum?id=TatRHT_1cK} {Quantifying memorization across neural language models}.
\newblock In \emph{The Eleventh International Conference on Learning Representations}.

\bibitem[{Carlini et~al.(2021)Carlini, Tramer, Wallace, Jagielski, Herbert-Voss, Lee, Roberts, Brown, Song, Erlingsson et~al.}]{carlini2021extracting}
Nicholas Carlini, Florian Tramer, Eric Wallace, Matthew Jagielski, Ariel Herbert-Voss, Katherine Lee, Adam Roberts, Tom~B Brown, Dawn Song, Ulfar Erlingsson, et~al. 2021.
\newblock \href {https://www.usenix.org/conference/usenixsecurity21/presentation/carlini-extracting} {Extracting training data from large language models.}
\newblock In \emph{USENIX Security Symposium}, volume~6.

\bibitem[{Chesterman(2023)}]{chesterman2023ai}
Simon Chesterman. 2023.
\newblock Ai-generated content is taking over the world. but who owns it?
\newblock \emph{But Who Owns it}.

\bibitem[{El-Mhamdi et~al.(2022)El-Mhamdi, Farhadkhani, Guerraoui, Gupta, Hoang, Pinot, and Stephan}]{el2022sok}
El-Mahdi El-Mhamdi, Sadegh Farhadkhani, Rachid Guerraoui, Nirupam Gupta, L{\^e}-Nguy{\^e}n Hoang, Rafael Pinot, and John Stephan. 2022.
\newblock \href {https://arxiv.org/abs/2209.15259} {On the impossible safety of large ai models}.
\newblock \emph{ArXiv preprint}, abs/2209.15259.

\bibitem[{Franceschelli and Musolesi(2023)}]{franceschelli2023creativity}
Giorgio Franceschelli and Mirco Musolesi. 2023.
\newblock \href {https://arxiv.org/abs/2304.00008} {On the creativity of large language models}.
\newblock \emph{ArXiv preprint}, abs/2304.00008.

\bibitem[{Frye(2022)}]{frye2022should}
Brian~L Frye. 2022.
\newblock Should using an ai text generator to produce academic writing be plagiarism?
\newblock \emph{Fordham Intellectual Property, Media \& Entertainment Law Journal, Forthcoming}.

\bibitem[{Gao et~al.(2022)Gao, Dai, Pasupat, Chen, Chaganty, Fan, Zhao, Lao, Lee, Juan et~al.}]{gao2022rarr}
Luyu Gao, Zhuyun Dai, Panupong Pasupat, Anthony Chen, Arun~Tejasvi Chaganty, Yicheng Fan, Vincent~Y Zhao, Ni~Lao, Hongrae Lee, Da-Cheng Juan, et~al. 2022.
\newblock \href {https://arxiv.org/abs/2210.08726} {Rarr: Researching and revising what language models say, using language models}.
\newblock \emph{ArXiv preprint}, abs/2210.08726.

\bibitem[{Gao et~al.(2023)Gao, Yen, Yu, and Chen}]{gao2023enabling}
Tianyu Gao, Howard Yen, Jiatong Yu, and Danqi Chen. 2023.
\newblock \href {https://arxiv.org/abs/2305.14627} {Enabling large language models to generate text with citations}.
\newblock \emph{ArXiv preprint}, abs/2305.14627.

\bibitem[{Greenberg(2009)}]{greenberg2009citation}
Steven~A Greenberg. 2009.
\newblock How citation distortions create unfounded authority: analysis of a citation network.
\newblock \emph{Bmj}, 339.

\bibitem[{Grosse et~al.(2023)Grosse, Bae, Anil, Elhage, Tamkin, Tajdini, Steiner, Li, Durmus, Perez et~al.}]{grosse2023studying}
Roger Grosse, Juhan Bae, Cem Anil, Nelson Elhage, Alex Tamkin, Amirhossein Tajdini, Benoit Steiner, Dustin Li, Esin Durmus, Ethan Perez, et~al. 2023.
\newblock Studying large language model generalization with influence functions.
\newblock \emph{arXiv preprint arXiv:2308.03296}.

\bibitem[{Guo et~al.(2022)Guo, Xie, Li, Lyu, and Zhang}]{guo2022threats}
Shangwei Guo, Chunlong Xie, Jiwei Li, Lingjuan Lyu, and Tianwei Zhang. 2022.
\newblock \href {https://arxiv.org/abs/2202.06862} {Threats to pre-trained language models: Survey and taxonomy}.
\newblock \emph{ArXiv preprint}, abs/2202.06862.

\bibitem[{Guu et~al.(2020)Guu, Lee, Tung, Pasupat, and Chang}]{pmlr-v119-guu20a}
Kelvin Guu, Kenton Lee, Zora Tung, Panupong Pasupat, and Ming{-}Wei Chang. 2020.
\newblock \href {http://proceedings.mlr.press/v119/guu20a.html} {Retrieval augmented language model pre-training}.
\newblock In \emph{Proceedings of the 37th International Conference on Machine Learning, {ICML} 2020, 13-18 July 2020, Virtual Event}, volume 119 of \emph{Proceedings of Machine Learning Research}, pages 3929--3938. {PMLR}.

\bibitem[{Honovich et~al.(2022)Honovich, Aharoni, Herzig, Taitelbaum, Kukliansy, Cohen, Scialom, Szpektor, Hassidim, and Matias}]{honovich2022true}
Or~Honovich, Roee Aharoni, Jonathan Herzig, Hagai Taitelbaum, Doron Kukliansy, Vered Cohen, Thomas Scialom, Idan Szpektor, Avinatan Hassidim, and Yossi Matias. 2022.
\newblock \href {https://doi.org/10.18653/v1/2022.dialdoc-1.19} {{TRUE}: Re-evaluating factual consistency evaluation}.
\newblock In \emph{Proceedings of the Second DialDoc Workshop on Document-grounded Dialogue and Conversational Question Answering}, pages 161--175, Dublin, Ireland. Association for Computational Linguistics.

\bibitem[{Huang and Chang(2023)}]{huang2022towards}
Jie Huang and Kevin Chen-Chuan Chang. 2023.
\newblock \href {https://doi.org/10.18653/v1/2023.findings-acl.67} {Towards reasoning in large language models: A survey}.
\newblock In \emph{Findings of the Association for Computational Linguistics: ACL 2023}, pages 1049--1065, Toronto, Canada. Association for Computational Linguistics.

\bibitem[{Huang et~al.(2023)Huang, Ping, Xu, Shoeybi, Chang, and Catanzaro}]{huang2023raven}
Jie Huang, Wei Ping, Peng Xu, Mohammad Shoeybi, Kevin Chen-Chuan Chang, and Bryan Catanzaro. 2023.
\newblock \href {https://arxiv.org/abs/2308.07922} {Raven: In-context learning with retrieval augmented encoder-decoder language models}.
\newblock \emph{arXiv preprint arXiv:2308.07922}.

\bibitem[{Huang et~al.(2022)Huang, Shao, and Chang}]{huang2022large}
Jie Huang, Hanyin Shao, and Kevin Chen-Chuan Chang. 2022.
\newblock \href {https://aclanthology.org/2022.findings-emnlp.148} {Are large pre-trained language models leaking your personal information?}
\newblock In \emph{Findings of the Association for Computational Linguistics: EMNLP 2022}, pages 2038--2047, Abu Dhabi, United Arab Emirates. Association for Computational Linguistics.

\bibitem[{Izacard and Grave(2021)}]{izacard-grave-2021-leveraging}
Gautier Izacard and Edouard Grave. 2021.
\newblock \href {https://doi.org/10.18653/v1/2021.eacl-main.74} {Leveraging passage retrieval with generative models for open domain question answering}.
\newblock In \emph{Proceedings of the 16th Conference of the European Chapter of the Association for Computational Linguistics: Main Volume}, pages 874--880, Online. Association for Computational Linguistics.

\bibitem[{Izacard et~al.(2022)Izacard, Lewis, Lomeli, Hosseini, Petroni, Schick, Dwivedi-Yu, Joulin, Riedel, and Grave}]{izacard2022few}
Gautier Izacard, Patrick Lewis, Maria Lomeli, Lucas Hosseini, Fabio Petroni, Timo Schick, Jane Dwivedi-Yu, Armand Joulin, Sebastian Riedel, and Edouard Grave. 2022.
\newblock \href {https://arxiv.org/abs/2208.03299} {Atlas: Few-shot learning with retrieval augmented language models}.
\newblock \emph{arXiv preprint arXiv}, 2208.

\bibitem[{Jannot et~al.(2013)Jannot, Agoritsas, Gayet-Ageron, and Perneger}]{jannot2013citation}
Anne-Sophie Jannot, Thomas Agoritsas, Ang{\`e}le Gayet-Ageron, and Thomas~V Perneger. 2013.
\newblock Citation bias favoring statistically significant studies was present in medical research.
\newblock \emph{Journal of clinical epidemiology}, 66(3):296--301.

\bibitem[{Ji et~al.(2023)Ji, Lee, Frieske, Yu, Su, Xu, Ishii, Bang, Madotto, and Fung}]{ji2023survey}
Ziwei Ji, Nayeon Lee, Rita Frieske, Tiezheng Yu, Dan Su, Yan Xu, Etsuko Ishii, Ye~Jin Bang, Andrea Madotto, and Pascale Fung. 2023.
\newblock Survey of hallucination in natural language generation.
\newblock \emph{ACM Computing Surveys}, 55(12):1--38.

\bibitem[{Koh and Liang(2017)}]{pmlr-v70-koh17a}
Pang~Wei Koh and Percy Liang. 2017.
\newblock \href {https://proceedings.mlr.press/v70/koh17a.html} {Understanding black-box predictions via influence functions}.
\newblock In \emph{Proceedings of the 34th International Conference on Machine Learning}, volume~70 of \emph{Proceedings of Machine Learning Research}, pages 1885--1894. PMLR.

\bibitem[{Lee et~al.(2023)Lee, Le, Chen, and Lee}]{lee2023language}
Jooyoung Lee, Thai Le, Jinghui Chen, and Dongwon Lee. 2023.
\newblock Do language models plagiarize?
\newblock In \emph{Proceedings of the ACM Web Conference 2023}, pages 3637--3647.

\bibitem[{Lewis et~al.(2020)Lewis, Perez, Piktus, Petroni, Karpukhin, Goyal, K{\"{u}}ttler, Lewis, Yih, Rockt{\"{a}}schel, Riedel, and Kiela}]{NEURIPS2020_6b493230}
Patrick S.~H. Lewis, Ethan Perez, Aleksandra Piktus, Fabio Petroni, Vladimir Karpukhin, Naman Goyal, Heinrich K{\"{u}}ttler, Mike Lewis, Wen{-}tau Yih, Tim Rockt{\"{a}}schel, Sebastian Riedel, and Douwe Kiela. 2020.
\newblock \href {https://proceedings.neurips.cc/paper/2020/hash/6b493230205f780e1bc26945df7481e5-Abstract.html} {Retrieval-augmented generation for knowledge-intensive {NLP} tasks}.
\newblock In \emph{Advances in Neural Information Processing Systems 33: Annual Conference on Neural Information Processing Systems 2020, NeurIPS 2020, December 6-12, 2020, virtual}.

\bibitem[{Li et~al.(2023)Li, Guo, Fan, Xu, Huang, Meng, and Song}]{li2023multi}
Haoran Li, Dadi Guo, Wei Fan, Mingshi Xu, Jie Huang, Fanpu Meng, and Yangqiu Song. 2023.
\newblock \href {https://doi.org/10.18653/v1/2023.findings-emnlp.272} {Multi-step jailbreaking privacy attacks on {C}hat{GPT}}.
\newblock In \emph{Findings of the Association for Computational Linguistics: EMNLP 2023}, pages 4138--4153, Singapore. Association for Computational Linguistics.

\bibitem[{Liu et~al.(2023)Liu, Zhang, and Liang}]{liu2023evaluating}
Nelson~F Liu, Tianyi Zhang, and Percy Liang. 2023.
\newblock \href {https://arxiv.org/abs/2304.09848} {Evaluating verifiability in generative search engines}.
\newblock \emph{ArXiv preprint}, abs/2304.09848.

\bibitem[{Menick et~al.(2022)Menick, Trebacz, Mikulik, Aslanides, Song, Chadwick, Glaese, Young, Campbell-Gillingham, Irving et~al.}]{menick2022teaching}
Jacob Menick, Maja Trebacz, Vladimir Mikulik, John Aslanides, Francis Song, Martin Chadwick, Mia Glaese, Susannah Young, Lucy Campbell-Gillingham, Geoffrey Irving, et~al. 2022.
\newblock \href {https://arxiv.org/abs/2203.11147} {Teaching language models to support answers with verified quotes}.
\newblock \emph{ArXiv preprint}, abs/2203.11147.

\bibitem[{Metzler et~al.(2021)Metzler, Tay, Bahri, and Najork}]{metzler2021rethinking}
Donald Metzler, Yi~Tay, Dara Bahri, and Marc Najork. 2021.
\newblock Rethinking search: making domain experts out of dilettantes.
\newblock In \emph{Acm sigir forum}, pages 1--27. ACM New York, NY, USA.

\bibitem[{OpenAI(2022)}]{openai2022chatgpt}
OpenAI. 2022.
\newblock Chatgpt: Optimizing language models for dialogue.
\newblock \emph{OpenAI}.

\bibitem[{OpenAI(2023)}]{openai2023gpt4}
OpenAI. 2023.
\newblock \href {http://arxiv.org/abs/2303.08774} {Gpt-4 technical report}.

\bibitem[{Page et~al.(1999)Page, Brin, Motwani, and Winograd}]{page1999pagerank}
Lawrence Page, Sergey Brin, Rajeev Motwani, and Terry Winograd. 1999.
\newblock The pagerank citation ranking: Bringing order to the web.
\newblock Technical report, Stanford InfoLab.

\bibitem[{Pan et~al.(2023)Pan, Pan, Chen, Nakov, Kan, and Wang}]{pan2023risk}
Yikang Pan, Liangming Pan, Wenhu Chen, Preslav Nakov, Min-Yen Kan, and William~Yang Wang. 2023.
\newblock \href {https://arxiv.org/abs/2305.13661} {On the risk of misinformation pollution with large language models}.
\newblock \emph{ArXiv preprint}, abs/2305.13661.

\bibitem[{Park et~al.(2023)Park, Georgiev, Ilyas, Leclerc, and Madry}]{park2023trak}
Sung~Min Park, Kristian Georgiev, Andrew Ilyas, Guillaume Leclerc, and Aleksander Madry. 2023.
\newblock Trak: Attributing model behavior at scale.
\newblock \emph{arXiv preprint arXiv:2303.14186}.

\bibitem[{Petroni et~al.(2019)Petroni, Rockt{\"a}schel, Riedel, Lewis, Bakhtin, Wu, and Miller}]{petroni2019language}
Fabio Petroni, Tim Rockt{\"a}schel, Sebastian Riedel, Patrick Lewis, Anton Bakhtin, Yuxiang Wu, and Alexander Miller. 2019.
\newblock \href {https://doi.org/10.18653/v1/D19-1250} {Language models as knowledge bases?}
\newblock In \emph{Proceedings of the 2019 Conference on Empirical Methods in Natural Language Processing and the 9th International Joint Conference on Natural Language Processing (EMNLP-IJCNLP)}, pages 2463--2473, Hong Kong, China. Association for Computational Linguistics.

\bibitem[{Rashkin et~al.(2023)Rashkin, Nikolaev, Lamm, Aroyo, Collins, Das, Petrov, Singh~Tomar, Turc, and Reitter}]{rashkin2023measuring}
Hannah Rashkin, Vitaly Nikolaev, Matthew Lamm, Lora Aroyo, Michael Collins, Dipanjan Das, Slav Petrov, Gaurav Singh~Tomar, Iulia Turc, and David Reitter. 2023.
\newblock Measuring attribution in natural language generation models.
\newblock \emph{Computational Linguistics}, pages 1--66.

\bibitem[{Shah and Bender(2022)}]{10.1145/3498366.3505816}
Chirag Shah and Emily~M. Bender. 2022.
\newblock \href {https://doi.org/10.1145/3498366.3505816} {Situating search}.
\newblock In \emph{Proceedings of the 2022 Conference on Human Information Interaction and Retrieval}, CHIIR '22, page 221–232, New York, NY, USA. Association for Computing Machinery.

\bibitem[{Shao et~al.(2024)Shao, Huang, Zheng, and Chang}]{shao2023quantifying}
Hanyin Shao, Jie Huang, Shen Zheng, and Kevin Chang. 2024.
\newblock \href {https://aclanthology.org/2024.findings-eacl.54} {Quantifying association capabilities of large language models and its implications on privacy leakage}.
\newblock In \emph{Findings of the Association for Computational Linguistics: EACL 2024}, pages 814--825, St. Julian{'}s, Malta. Association for Computational Linguistics.

\bibitem[{Shi et~al.(2023)Shi, Min, Yasunaga, Seo, James, Lewis, Zettlemoyer, and Yih}]{shi2023replug}
Weijia Shi, Sewon Min, Michihiro Yasunaga, Minjoon Seo, Rich James, Mike Lewis, Luke Zettlemoyer, and Wen-tau Yih. 2023.
\newblock \href {https://arxiv.org/abs/2301.12652} {Replug: Retrieval-augmented black-box language models}.
\newblock \emph{ArXiv preprint}, abs/2301.12652.

\bibitem[{Taylor et~al.(2022)Taylor, Kardas, Cucurull, Scialom, Hartshorn, Saravia, Poulton, Kerkez, and Stojnic}]{taylor2022galactica}
Ross Taylor, Marcin Kardas, Guillem Cucurull, Thomas Scialom, Anthony Hartshorn, Elvis Saravia, Andrew Poulton, Viktor Kerkez, and Robert Stojnic. 2022.
\newblock \href {https://arxiv.org/pdf/2211.09085.pdf} {Galactica: A large language model for science}.
\newblock \emph{ArXiv preprint}, abs/2211.09085.

\bibitem[{Vaswani et~al.(2017)Vaswani, Shazeer, Parmar, Uszkoreit, Jones, Gomez, Kaiser, and Polosukhin}]{vaswani2017attention}
Ashish Vaswani, Noam Shazeer, Niki Parmar, Jakob Uszkoreit, Llion Jones, Aidan~N. Gomez, Lukasz Kaiser, and Illia Polosukhin. 2017.
\newblock \href {https://proceedings.neurips.cc/paper/2017/hash/3f5ee243547dee91fbd053c1c4a845aa-Abstract.html} {Attention is all you need}.
\newblock In \emph{Advances in Neural Information Processing Systems 30: Annual Conference on Neural Information Processing Systems 2017, December 4-9, 2017, Long Beach, CA, {USA}}, pages 5998--6008.

\bibitem[{Wang et~al.(2023)Wang, Ping, Xu, McAfee, Liu, Shoeybi, Dong, Kuchaiev, Li, Xiao, Anandkumar, and Catanzaro}]{wang2023shall}
Boxin Wang, Wei Ping, Peng Xu, Lawrence McAfee, Zihan Liu, Mohammad Shoeybi, Yi~Dong, Oleksii Kuchaiev, Bo~Li, Chaowei Xiao, Anima Anandkumar, and Bryan Catanzaro. 2023.
\newblock \href {https://doi.org/10.18653/v1/2023.emnlp-main.482} {Shall we pretrain autoregressive language models with retrieval? a comprehensive study}.
\newblock In \emph{Proceedings of the 2023 Conference on Empirical Methods in Natural Language Processing}, pages 7763--7786, Singapore. Association for Computational Linguistics.

\bibitem[{Wei et~al.(2022)Wei, Wang, Schuurmans, Bosma, brian ichter, Xia, Chi, Le, and Zhou}]{wei2022chain}
Jason Wei, Xuezhi Wang, Dale Schuurmans, Maarten Bosma, brian ichter, Fei Xia, Ed~H. Chi, Quoc~V Le, and Denny Zhou. 2022.
\newblock \href {https://openreview.net/forum?id=_VjQlMeSB_J} {Chain of thought prompting elicits reasoning in large language models}.
\newblock In \emph{Advances in Neural Information Processing Systems}.

\bibitem[{Yue et~al.(2023)Yue, Wang, Chen, Zhang, Su, and Sun}]{yue2023automatic}
Xiang Yue, Boshi Wang, Ziru Chen, Kai Zhang, Yu~Su, and Huan Sun. 2023.
\newblock \href {https://doi.org/10.18653/v1/2023.findings-emnlp.307} {Automatic evaluation of attribution by large language models}.
\newblock In \emph{Findings of the Association for Computational Linguistics: EMNLP 2023}, pages 4615--4635, Singapore. Association for Computational Linguistics.

\bibitem[{Zheng et~al.(2023)Zheng, Huang, and Chang}]{zheng2023does}
Shen Zheng, Jie Huang, and Kevin Chen-Chuan Chang. 2023.
\newblock \href {https://openreview.net/forum?id=w7o14LCw9P} {Why does {ChatGPT} fall short in providing truthful answers?}
\newblock In \emph{I Can't Believe It's Not Better Workshop: Failure Modes in the Age of Foundation Models}.

\end{thebibliography}

\end{document}